\documentclass[11pt,a4paper]{article}

 \usepackage[round]{natbib}
\usepackage{graphicx}
\usepackage{booktabs}
\usepackage{multirow}
\usepackage[table]{xcolor}
\usepackage[hyperref]{acl2020}
\usepackage{times}
\usepackage{latexsym}

\usepackage{microtype}

\aclfinalcopy 

\title{Rating Facts under Coarse-to-fine Regimes}

\author{Guojun Wu \\
  National Taiwan University of Science and Technology \\
  
  \texttt{b10702127@ntust.edu.tw} \\\
  }

\date{}

\begin{document}
\maketitle
\begin{abstract}
The rise of manipulating fake news as a political weapon has become a global concern and highlighted the incapability of manually fact checking against rapidly produced fake news. Thus, statistical approaches are required if we are to address this problem efficiently. The shortage of publicly available datasets is one major bottleneck of automated fact checking. To remedy this,                                         we collected 24K manually rated statements from PolitiFact\footnote{https://www.politifact.com/}. The class values exhibit a natural order with respect to truthfulness as shown in Table~\ref{meter}. Thus, our task represents a twist from standard classification, due to the various degrees of similarity between classes. To investigate this, we defined coarse-to-fine classification regimes, which presents new challenge for classification. To address this, we propose BERT-based models. After training, class similarity is sensible over the multi-class datasets, especially in the fine-grained one. Under all the regimes, BERT achieves state of the art, while the additional layers provide insignificant improvement.
\end{abstract}

\section{Introduction}

The manipulation of weaponized fake news has drawn public concern in many political events. One study about fake news \citep{10.1257/jep.31.2.211} reported that average American adult encountered one or several fake news stories during 2016 election period. Since their database was limited, this estimate can be conservative.  Moreover, the steep decline of public trust in mass media in 2016 \citep{unknown-author-2021} also drew attention to the impact of fake news. 

To limit the widespread of fake news, many fact-checking websites have emerged such as PolitiFact and First Draft\footnote{https://firstdraftnews.org/}. Take PolitiFact, a reporter will research on the statement and provide detailed analysis. Then several editors will rate the truthfulness together. While this fact-checking procedure is responsible, it will consume a lot effort and time. 

Automatic fact checking seems to be an elegant approach to detect fake news quickly and cut the spread. However, verification of novel claims is challenging because of the difficulty to retrieve relative information. Thus, we expect to assess the truthfulness of a claim purely depending on linguistic analysis.\citet{wang2017liar} has discussed the lack of available datasets about fake news, which limits the potential to combat fake news. To address this, they introduced a new benchmark dataset, LIAR, which contains 12.8K rated statements from PolitiFact. In our work, we constructed a similar dataset relatively larger than LIAR.

\begin{table}
\centering
\begin{tabular}{p{7cm}}

\rowcolor {green!90!yellow!50} \textbf{True:} The statement is accurate and there is nothing significant missing.  \\

\rowcolor{green!70!yellow!40} \textbf{Mostly true:} The statement is accurate but needs clarification or additional information.  \\ 

\rowcolor{green!60!yellow!30} \textbf{Half true:} The statement is partially accurate but leaves out important details or takes thing out of context.  \\ 

\rowcolor{red!0!yellow!100} \textbf{Mostly false:} The statement contains an element of truth but ignores critical fact that would give a different impression.  \\ 

\rowcolor{red!25!yellow!100} \textbf{False:} The statement is not accurate.  \\ 

\rowcolor{red!40!yellow!100} \textbf{Pants on fire: } The statement is not accurate and makes a ridiculous claim.   \\ 

\end{tabular}
\caption{\label{meter} Meter of truthfulness rating. }
\end{table}

Empirically, our task is about rating-inference, which represents some difference from standard multi-class text classification. The various degrees of similarity between classes can be confusing for classification. One study \citep{pang-lee-2005-seeing} suggests that as the class similarity increases, performance of both human and models drop obviously. This problem has also been discussed in prior work about fact checking \citep{vlachos-riedel-2014-fact,rashkin2017truth}. In order to investigate this, we defined three classification regimes (i.e., fine-grained, coarse-grained, and binary -- 6-class, 3-class, and 2-class respectively). 

In this paper, we propose BERT-based models to rate the statements. Pre-trained language models have been a significant ingredient in multiple NLP tasks. While the standard BERT has achieved state of the art \citep{devlin-etal-2019-bert}, we hypothesis that the approach for classification representation can be enhanced. We propose a structure with the recurrent (i.e., BiLSTM \citep{sachan2019revisiting}) or convolutional (i.e., CNN \citep{kim-2014-convolutional}) layers on top of BERT. Both the additional layers have achieved competitive performance in many text classification tasks based on word vectors. For comparison, we use BERT as our baseline model.

Our experiments suggest that class similarity can influence the judgement of the model. This situation is more obvious in fine-grained dataset. Moreover, BERT achieves state-of-the-art performance under all three regimes. While the additional BiLSTM fails to advance the results, the CNN consistently provides slight improvement.

\section{Related work}

\paragraph{\textbf{Datasets}} A prior work \citep{vlachos-riedel-2014-fact} provides one of the first fact-checking datasets, which consists of 106 claims from PolitiFact. Later, \citet{wang2017liar} introduce a new benchmark dataset, LIAR. It includes 12.8K claims also from PolitiFact and their meta-data (e.g., the context of the claim). More recent work \citep{rashkin2017truth} has collected 10,483 claims from PolitiFact and its spin-off sites. Our paper introduces datasets relatively larger than these datasets of similar type, and digs deeper about classification regimes.
\paragraph{\textbf{Approaches}} There have been various kinds of linguistic analysis, such as separation of fake news types \citep{rubin2015deception}, article structure and content of hoaxes \citep{kumar2016disinformation}, linguistic style of clickbaits  \citep{biyani20168}, quantitative study of linguistic differences \citep{rashkin2017truth}, and bias of language patterns in fake news \citep{o2018language}. Except for these, claim verification has often been related to stance classification. Claims and associated articles are required in the datasets, such as Emergent \citep{ferreira2016emergent}  and FEVER \citep{thorne-etal-2018-fever}. The claim-level truthfulness assessment can be reached through the article-level stance classification \citep{mohtarami-etal-2018-automatic,xu2019adversarial,baly-etal-2018-integrating}. While stance classification can detect repetition or paraphrase of existed claims, it cannot deal with the novel ones. Thus, our work takes the linguistic approach, utilizing deep learning to assess the claims.

\section{Politic Fact Dataset}

In this section, we introduce and provide some analyses for the new Politic Fact datasets, which include labeled statements collected from PolitiFact. As shown in Table~\ref{meter}, there are various degrees of similarity between classes; for example, “True” is closer to “Mostly true” than to “false”. To further investigate this, we defined coarse-to-fine classification regimes. Accordingly, we constructed three datasets with different regimes. To make the classes balanced, we kept all statements in rare class and randomly selected equal number of statements in abundant class. Due to this filtering, the size of each dataset can vary. We report the specifications of these datasets in Table~\ref{spec}.

\newcommand{\ra}[1]{\renewcommand{\arraystretch}{#1}}
\begin{table}
\centering
\ra{1.3}
\begin{tabular}{crrc}
\toprule
\textbf{Regime} & \textbf{Train} & \textbf{Test} & \textbf{Class} \\ 
\midrule
Fine-grained & 11932 & 2987 & 6 \\
Coarse-grained & 16980 & 4245 & 3 \\
Binary & 11320 & 2830 & 2\\
\bottomrule
\end{tabular}
\caption{\label{spec} Specifications of Politic Fact datasets. }
\end{table}

\subsection{FPF: Fine-grained Politic Fact}

In this dataset, we keep the original labels (6-class) in PolitiFact, as shown in Table~\ref{meter}. Since it is meticulously classified, we will name this Fine-grained Politic Fact (FPF). This regime can capture the main variability of statements, with well-balanced labels from “True” to “Pants on fire”. 

When we began to construct this 6-class dataset, we found that the most false class (a.k.a., “Pants on fire”) contained only about 10\% of all the statements. Thus, after filtering about 40\% of the data, this dataset included 14.9K statements. Then all the statements were split into train (11932) and test split (2984). 

\subsection{CPF: Coarse-grained Politic Fact}

One problem with the FPF is that the high similarity between close classes makes classification challenging. For example, the difference between “True” and “Mostly true” is hard to tell. Since both of them indicate that the statement is accurate, while the latter one needs additional information.  

To address these, we proposed a new dataset -- Coarse-grained Politic Fact (CPF). We treated top two truthfulness ratings as true, the middle two as neutral and the lowest two as false. This dataset shrinked the number of classes to three and reduced the similarity between classes. Then, we filtered and split the dataset into train (16980) and test split (4245).

\subsection{BPF: Binary Politic Fact}

The simplest member of this family of Politic Fact dataset is the Binary Politic Fact (BPF). This dataset consists of either true or false class. We constructed this based on CPF, ignoring the neutral class. This filtered 1/3 of CPF with the two sets (train/test) having 11320/2830 statements.

\section{Models}

The models in this section all base on BERT \citep{devlin-etal-2019-bert}. BERT utilizes the [CLS] token as the representation and has already achieved state-of-the-art results in many text classification tasks, such as SST-2 \citep{socher-etal-2013-recursive}. However, we believe that with the additional layers the model can capture the sentence representation better.

\subsection{BERT}
Following the standard way \citep{devlin-etal-2019-bert}, we take the [CLS] token of the final layer as the sentence representation. This model is the baseline model of our task.

\subsection{BERT-BiLSTM}
Since BERT has proven capable to capture high-level feature, we take advantage of it by outputting the entire final layer. Then we make use of the BiLSTM \citep{sachan2019revisiting}---We first pass the output to a BiLSTM layer. Next, we concatenate the forward LSTM and backward LSTM at each time step. Then we apply max-over-time operation over the concatenated hidden states to obtain the sentence representation.

\subsection{BERT-CNN}
In this model, we utilize CNN \citep{kim-2014-convolutional} for sentence representation. We first build the sentence matrix with the final layer of BERT. Then we perform one-dimension convolution on it to produce the feature maps with multiple filters. Then we run a max-over-time pooling over each feature maps to capture the most significant features. Next, we concatenated these features to form the sentence representation. 

\section{Experiment}
In this section, we present our analysis of class relationship of the data, compare performance of our models, and discuss about the necessity to design a new metric.

\subsection{Hyperparameters}
\paragraph{\textbf{BERT}} We made use of the 12-layer BERT-base. When we searched parameters for BERT, we followed the fine-tuning procedure \citep{devlin-etal-2019-bert}. The dropout rate was always 0.1. After an exhaustive search with the recommended parameters, we found that when batch size was 32, learning rate (Adam) was 5e-5, and the number of epochs was 4, the model performed the best.

\paragraph{\textbf{BERT with additional layers}} we kept the parameters in BERT the same. In BiLSTM, we kept hidden size the same as the dimension of the token (768) in BERT. Dropout rate was 0.5, following \citet{sachan2019revisiting}. In CNN, we follow the sensitivity analysis of CNN \citep{zhang2015sensitivity}. After comparing the results of several suggested settings. The model performed best when region size of filters was (7,7,7,7), the number of feature maps was 768 and the dropout rate was 0.5. 

\paragraph{\textbf{Fine tuning}} These parameters were chosen after search on a more true/false dataset (i.e., top three truthfulness ratings as more true and the others as more false). We do not otherwise perform any tuning on the three Politic Fact datasets.

\begin{figure}[t]
\includegraphics[width=7cm]{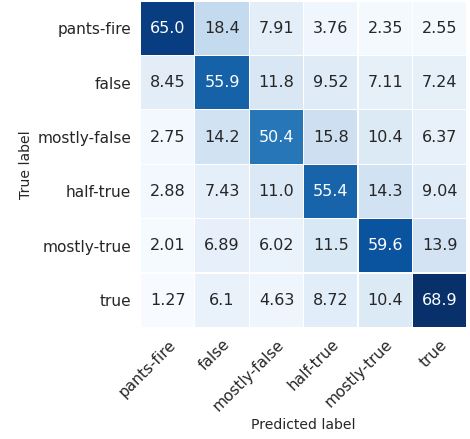}
\centering
\caption{\label{d1}Normalized distribution of predictions over FPF. (Average of three random seeds)}
\end{figure}

\subsection{Result analysis: Class similarity}
\begin{figure}[t]
\includegraphics[width=7cm]{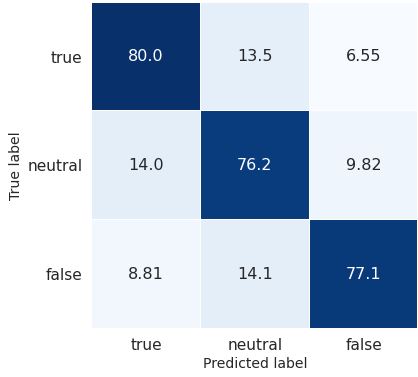}
\centering
\caption{\label{d2}Normalized distribution of predictions over CPF. (Average of three random seeds)}
\end{figure}

We present the distribution of predictions by BERT over FPF (6-class) and CPF (3-class) test set in Figure ~\ref{d1} and Figure~\ref{d2} respectively. The proportion of predicted labels decrease as the distance from the ground truth become further. Moreover, the polarized classes present higher accuracy than the neutral ones. These situations indicate that class similarity can fool our model to choose the classes that are close to the ground truth. Especially for neutral classes, the statements are more likely to be misclassified because they locate in the middle and have close relationship with both sides. While the model classify the statements more accurately in CPF, class similarity can still have slight impact on the judgement.

\subsection{Result analysis: Classifier comparison}
Table~\ref{comp} shows our models performance over three datasets. We chose weighted averaged F1 score as the metric to evaluate the performance. Since we utilized various random seeds, we report the average of five random seeds, with standard deviation as subscripts. Our baseline model (i.e., BERT) already performs well on its own. We had hoped to have performance gains through an additional BiLSTM layer. However, it only outperforms the baseline slightly on CPF.  Moreover, BERT-CNN consistently outperforms the other two models over all three datasets. 

\subsection{Metric}
In this section, we discuss about a new evaluation metric and we hope it can throw some light on further works. Intuitively, rating the ground truth as closer classes should be less wrong than as further classes. However, widely used evaluation metrics (i.e., accuracy and F1 score) only evaluate whether the model can hit the target (a.k.a., ground truth) which makes it unable to fully represent rating performance. To address this, we suggest mean absolute error (MAE) as a new metric. It can differentiate the scores through the gaps between the predictions and the target. Further research is needed on how to quantify the gaps appropriately.

\begin{table}
\centering
\ra{1.3}
\begin{tabular}{cccc}
\toprule
\textbf{Model} & \textbf{FPF} & \textbf{CPF} & \textbf{BPF} \\ 
\midrule
BERT & $61.4_{0.4}$ & $78.0_{0.4}$ & $87.4_{0.1}$ \\
BERT-BiLSTM & $61.4_{0.6}$ & $\mathbf{78.2_{0.3}}$ & $87.4_{0.3}$ \\
BERT-CNN & $\mathbf{61.8_{0.6}}$ &  $\mathbf{78.2_{0.3}}$ &  $\mathbf{87.8_{0.3}}$ \\
\bottomrule
\end{tabular}
\caption{\label{comp} F1 scores for Politic Fact datasets. We represent averages of five random seeds. Standard deviations are shown as subscripts. }
\end{table}

\section{Conclution}
In this paper, we introduce the Politic Fact datasets with coarse-to-fine regimes for fact-checking researches. Through analysis on the distribution of predictions, we have shown that class similarity is sensible in both fine-grained and coarse-grained datasets. Though the additional layers cannot provide significant improvement, BERT is capable to tackle the task. In order to address the limit of current metrics, we also suggest a new metric and hope it inspire further researches.

\bibliographystyle{acl_natbib}
\bibliography{acl2020}

\end{document}